\pdfoutput=1

\documentclass[11pt]{article}

\usepackage[preprint]{acl}
\usepackage{xspace}
\usepackage{booktabs}
\usepackage{bbding}
\usepackage{arydshln}
\usepackage{times}
\usepackage{latexsym}
\usepackage{tcolorbox}

\usepackage[T1]{fontenc}

\usepackage[utf8]{inputenc}

\usepackage{microtype}

\usepackage{inconsolata}

\usepackage{graphicx}

\usepackage[utf8]{inputenc} 
\usepackage[T1]{fontenc}    
\usepackage{hyperref}       
\usepackage{url}            
\usepackage{booktabs}       
\usepackage{amsfonts}       
\usepackage{nicefrac}       
\usepackage{microtype}      
\usepackage{xcolor}         
\usepackage{multirow,multicol}
\usepackage{graphicx}

\usepackage{enumitem}

\title{Tuning LLMs by RAG Principles: Towards LLM-native Memory}

\author{
  Jiale Wei$\quad$ Shuchi Wu$\quad$ Ruochen Liu$\quad$ Xiang Ying$\quad$ Jingbo Shang\thanks{$\ $  Corresponding author.}$\quad$ Fangbo Tao\\
  \\
  \{yingxiang, tao\}@mindverse.ai \\
  \\
  Mindverse.ai
}

\usepackage{xspace}

\newcommand{\our}{RAG-Tuned-LLM\xspace}

\begin{document}
\maketitle
\begin{abstract}
    Memory, additional information beyond the training of large language models (LLMs), is crucial to various real-world applications, such as personal assistant.
The two mainstream solutions to incorporate memory into the generation process are long-context LLMs and retrieval-augmented generation (RAG).
In this paper, we first systematically compare these two types of solutions on three renovated/new datasets and show that
(1) long-context solutions, although more expensive, shall be easier to capture the big picture and better answer queries which require considering the memory as a whole; 
and (2) when the queries concern specific information, RAG solutions shall be more competitive especially when the keywords can be explicitly matched.
Therefore, we propose a novel method \our which fine-tunes a relative small (e.g., 7B) LLM using the data generated following the RAG principles, so it can combine the advantages of both solutions.
Extensive experiments on three datasets demonstrate that \our can beat long-context LLMs and RAG methods across a wide range of query types.

\end{abstract}

\section{Introduction}

\begin{figure*}[t]
    \centering
    \includegraphics[width=\textwidth]{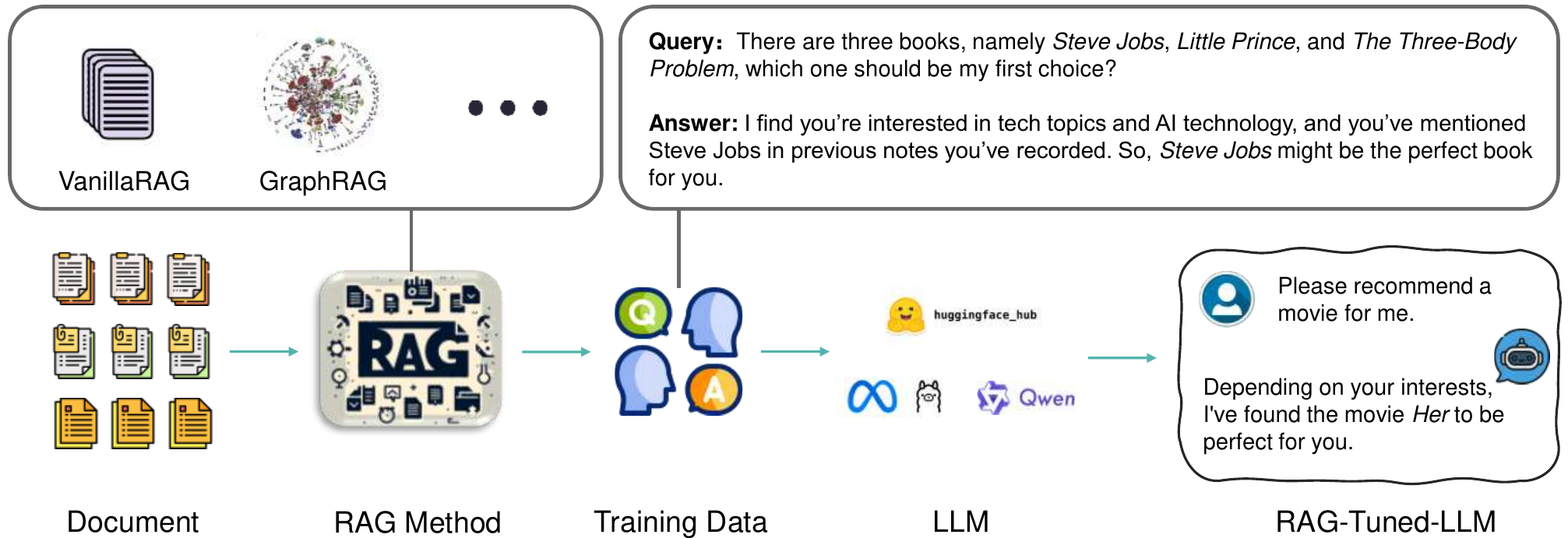}
    \caption{Overview of our \our method. \textbf{Stage 1}: RAG provides the foundation for synthesizing training data (query-answer pairs) for fine-tuning. \textbf{Stage 2}: The synthesized data is used to fine-tune a large language model (LLM) via LoRA. \textbf{Stage 3}: Inference is performed exclusively with LLM-native memory, eliminating the need for external memory. The \our combines the strengths of LLM-native solutions and RAG methods.}
    \label{fig:pipeline}
\end{figure*}

Memory, additional information beyond the training of large language models (LLMs), is crucial to various real-world applications, such as personal assistant~\citep{Mai2023LLMAA}.
The most intuitive solution to enable long memory into the generation process is long-context LLM, for example, 128K-token GPT-4o~\citep{achiam2023gpt}, 1M-or 10M-token Gemini 1.5~\citep{reid2024gemini}, or an LLM with ``unlimited'' context lengths by length extrapolation~\citep{peng2023yarn,xiao2023efficient,han2023lm,zhang2024soaring} and position bias~\citep{liu2024lost,peysakhovich2023attention,an2024make}. 
Retrieval-augmented generation (RAG)~\cite{lewis2020retrieval,kovcisky2018narrativeqa,pang2022quality,trivedi2022musique,edge2024local} is another popular approach to incorporate memory in a plug-in manner: a retriever identifies a small number of query-relevant contexts from a large corpus, and then feeds them into an LLM to answer the query. 
Compared with long-context LLMs, RAG's serving cost is more affordable, and therefore, RAG is potentially more popular than long-context LLMs in real-world applications. 

In this paper, we first systematically compare these two types of methods on three renovated/new datasets.
We start with two public datasets, namely \textit{news} articles~\cite{tang2024multihop} and \textit{podcast} transcripts~\cite{podcast}, following the general ideas mentioned in~\citet{edge2024local} to generate queries and references.
On these two datasets, we use the entire corpus as the memory.
We categorize the queries into two types, \emph{local} and \emph{global}.
Specifically, local queries target specific information and concrete answers from small chunks of memory. 
Global queries, on the other hand, require considering memory as a whole to generate high-level answers.
We further introduce a new proprietary dataset containing \emph{journaling} articles and user-provided \emph{local}/\emph{global} queries and their expected answers from our journaling app\footnote{Me.bot: \href{https://app.me.bot/}{https://app.me.bot/}}.

Intuitively, (1) long-context solutions, although more expensive, shall be easier to capture the big picture and better answer \emph{global} queries; 
and (2) when the queries concern \emph{local} information, RAG solutions shall be more competitive especially when the keywords can be explicitly matched. 
Based on these three datasets, we run competitions between a vanilla RAG~\cite{lewis2020retrieval} and Gemini 1.5~\cite{reid2024gemini}, with the win rate results shown in Table~\ref{long_context_rag}, confirming our intuitions. 
It is worth mentioning that RAG surpasses long-context LLMs when handling \emph{local} queries, yet underperforms in addressing \emph{global} ones.

Following our findings, we propose a novel LLM-native method \our which fine-tunes a relatively small (e.g., 7B) LLM using the data generated following the RAG principles, so it can combine the advantages of RAG and long-context solutions.
We call it \emph{LLM-native} because it maintains the same speed as directly prompting an LLM with only the question --— without requiring long contexts or retrieval from a knowledge base.
It enables the LLM to parameterize knowledge in a way that allows it to maintain contextual coherence and handle different types of queries more naturally and efficiently.
Specifically, as illustrated in Figure~\ref{fig:pipeline}, we follow the GraphRAG~\cite{edge2024local} principles to extract useful information from plain text documents. 
We then generate data from both \emph{local} and \emph{global} perspectives:
(1) \emph{local} data synthesis concentrates on generating content-specific query and answer pairs, 
and (2) \emph{global} data synthesis focuses on producing query-answer pairs that integrate insights across entities and relationships.
And finally, we employ the widely adopted LoRA technique~\cite{hu2021lora} to fine-tune the LLM.

Our experiments then demonstrate that \our can beat long-context LLMs and RAG methods on both {local} and {global} queries, and further case studies show that \our excels in providing insightful and user-friendly responses. Our codes have been released to the public at Github\footnote{https://github.com/mindverse/rag-tuned-llm}.

Our contributions are summarized as follows.
\begin{itemize}[nosep,leftmargin=*]
    \item We create three datasets with \emph{local} and \emph{global} queries with their references, and then systematically compare \emph{LLM-native} and (vanilla) RAG solutions, showing their respective unique advantages.
    It is worth mentioning that on one dataset, both the queries and references are manually created by human users.
    \item We follow the comparison results and propose a novel LLM-native method \our to combine the advantages of RAG and long-context solutions.
    \item Extensive experiments on three datasets demonstrate that \our can indeed outperform long-context LLMs and (advanced) RAG methods on both \emph{local} and \emph{global} queries.
\end{itemize}

\section{Long-context vs. RAG}

To motivate our work, we systematically compare long-context and RAG solutions and discuss their respect strengths in this section. 

\subsection{Settings}
\paragraph{Datasets.}
We consider three datasets for comparison, as detailed in Table~\ref{tbl:dataset_stats}. For the two public datasets—\textbf{News} articles~\cite{tang2024multihop} and \textbf{Podcast} transcripts~\cite{podcast}—we follow ~\citet{edge2024local} to generate 125 \emph{local} queries and 125 \emph{global} queries for each, along with their corresponding references. The \textbf{Journaling} dataset, newly introduced by us, is proprietary and derived from our journaling app. It contains 45 \emph{local} queries and 15 \emph{global} queries designed by users, accompanied by their expected answers. Users were informed to craft queries aimed at complex and nuanced scenarios, prioritizing reasoning capabilities over simple retrieval.
It is designed to robustly evaluate models' ability to handle intricate reasoning tasks in diverse real-world scenarios. It extends beyond basic fact retrieval to assess how well models can retrieve specific details while performing higher-order reasoning. Please refer to Table \ref{tbl:dataset_stats} for detailed statistics.

\paragraph{Methods.}
For the long-context LLM, we choose \texttt{Gemini-1.5-pro-001} due to its remarkable 2-million-token context window, which stands out as one of the longest among widely recognized and authoritative LLMs. This extensive context capacity sufficiently accommodates our experimental needs without requiring truncation.
For the RAG methods, we implement \texttt{VanillaRAG} using standard embedding and reranking techniques from the Langchain framework \footnote{\href{https://www.langchain.com/}{LangChain: https://www.langchain.com/}}. 
Specifically, VanillaRAG employs the \texttt{text-embedding-ada-002} model for initial chunk retrieval, selecting the top-$10$ most relevant chunks. 
These chunks are then refined using Cohere’s \texttt{rerank-english-v3.0} model, which filters the 10 chunks down to 3. 
We use \texttt{GPT-4o-mini}\footnote{Our small-scale experiment shows that \texttt{GPT-4o-mini} as the language model for answer generation in VanillaRAG delivers comparable performance with significantly lower cost than \texttt{GPT-4o}.} considering the cost efficiency and performance. 
By incorporating both embedding-based recall and reranking, this method serves as a strong RAG solution.

\begin{table}[t]
\centering
\caption{Dataset statistics. \textbf{Memory refers to the raw texts that will be utilized as additional information for answering queries.} Evaluation queries are split into \emph{local} and \emph{global} partitions according to their scopes. 
}
\label{tbl:dataset_stats}
\small
\resizebox{\linewidth}{!}{
\begin{tabular}{c cc ccc}
\toprule
\multirow{2}{*}{\textbf{Dataset}} & \multicolumn{2}{c}{\textbf{Memory}} & \multicolumn{3}{c}{\textbf{Evaluation Queries}}  \\ 
\cmidrule(lr){2-3} \cmidrule(lr){4-6}
                         & \# Docs & \# Tokens      & Global & Local & Avg Tokens \\ \midrule
Podcast                  & 66   & 832K &  125    & 125   & 22.30  \\
News                     & 609  & 1214K & 125    & 125   & 22.02  \\
Journaling               & 538   & 230K & 45     & 15    & 39.57  \\
\bottomrule
\end{tabular}
}
\end{table}

\begin{table}[t]
\centering
\caption{Wining rates of \textbf{Gemini-1.5} over \textbf{VanillaRAG} on \emph{local} and \emph{global} queries across three datasets using the four introduced metrics. Values exceeding \textbf{50\%} indicate that Gemini-1.5 outperforms VanillaRAG.}
\small
\resizebox{\linewidth}{!}{
\begin{tabular}{@{}clccc@{}}
\toprule
\textbf{Dataset}                  & \textbf{Metric}            & \textbf{Local} & \textbf{Global} & \textbf{Overall} \\ \midrule
\multirow{5}{*}{Podcast} & Helpful        & 81.60\% & 86.40\%  & 84.00\%   \\
                         & Rich          & 87.20\% & 90.40\%  & 88.80\%   \\
                         & Insightful   & 90.40\% & 90.40\%  & 90.40\%   \\
                         & User-Friendly & 85.60\% & 88.80\%  & 87.20\%   \\ \cmidrule{2-5}
                         & Overall & 86.20\% & 89.00\%  & 87.60\% \\ \midrule
\multirow{5}{*}{News}    & Helpful        & 46.40\% & 56.60\%  & 51.20\%   \\
                         & Rich          & 48.80\% & 56.80\%  & 52.80\%   \\
                         & Insightful    & 49.60\% & 58.40\%  & 54.00\%   \\
                         & User-Friendly & 46.40\% & 58.40\%  & 52.40\%   \\ \cmidrule{2-5}
                         & Overall & 47.80\% & 57.55\%  & 52.60\% \\ \midrule
\multirow{5}{*}{Journaling} & Helpful & 53.33\% & 93.33\% & 83.33\% \\
                         & Rich          & 46.67\% & 88.80\%  & 80.00\%   \\
                         & Insightful    & 53.33\% & 91.11\%  & 81.67\%   \\
                         & User-Friendly & 53.33\% & 93.33\%  & 83.33\%   \\ \cmidrule{2-5}
                         & Overall & 51.67\% & 91.64\%  & 82.08\% \\
\bottomrule
\end{tabular}
}
\label{long_context_rag}
\end{table}

\begin{table*}[t]
\centering
\caption{Graph statistics for the three datasets. The Graph Statics columns summarize the number of extracted entities, relations, and communities. The Synthesized SFT Data columns detail the number of generated queries, average query token count, and average answer token count.}
\label{tbl:graph_rag_stats}
\small
\begin{tabular}{c ccc ccc}
\toprule
 & \multicolumn{3}{c}{\textbf{Graph Statistics}} & \multicolumn{3}{c}{\textbf{Our Synthesized SFT Data}} \\ 
 \cmidrule(lr){2-4} \cmidrule(lr){5-7}
\textbf{Dataset}          & \textbf{Entities} & \textbf{Relations} & \textbf{Communities} & \textbf{\# of Queries} & \textbf{Avg Query Tokens} & \textbf{Avg Answer Tokens}\\
\midrule
Podcast          & 5,182    & 8,631         & 837    &  54,627   & 23.29        & 264.04 \\
News             & 17,877   & 26,208        & 3,534   & 155,896  & 23.54        & 273.19   \\
Journaling & 2,930    & 3,751         & 547      &  18,355  & 36.46        & 562.60 \\
\bottomrule
\end{tabular}
\end{table*}

\subsection{Evaluation Metrics}
We design our evaluation criteria to ensure that the generated answers are not only accurate but also practically helpful for real-world applications, such as personal assistants. We refer to the attribute perspectives in ~\citep{Li2024FromGT} and ranking prioritization in ~\citep{Wang2024HelpSteer2PreferenceCR} as:
\begin{itemize}[nosep, leftmargin=*]
    \item \textbf{Helpful} assesses the precision, contextual relevance, and practical value of the response in effectively addressing the query.
    \item \textbf{Rich} measures the comprehensiveness, depth, and diversity of perspectives of the response.
    \item \textbf{Insightful} evaluates the profundity of understanding and the uniqueness of insights offered.
    \item \textbf{User-Friendly} focuses on the clarity, coherence, and accessibility of the response.
\end{itemize}
In Table \ref{long_context_rag}, we additionally report an ``overall'' metric, calculated as the average performance across the aforementioned four metrics. More detailed explanations of these metrics are deferred to Appendix~\ref{appendix: metric}.

We evaluate responses from two competitors on various queries and compute the winning rate of one method over the other. We adopt an LLM as the judge, comparing the two answers based on the target metric, the query, and a reference answer. The reference answer, meticulously crafted and verified, provides a solid foundation for the LLM's comparison. To mitigate stochastic variability, this evaluation process is repeated multiple times. Notably, in our experiments, we observed comparable judging performance between \texttt{GPT-4o-mini} and \texttt{GPT-4o}. For cost efficiency, we report results using \texttt{GPT-4o-mini}. After aligning the LLM’s evaluations with human assessments, we found a concordance rate of 86\%, which is high enough for fair comparison, with 215 out of 250 cases exhibiting agreement. Considering the cost and insights from GraphRAG, we believe that the size of this test set is quite convincing.

\subsection{Results}
We present the winning rates of the long-context LLM compared to VanillaRAG in Table \ref{long_context_rag}. The data reveals that the long-context solution, though more expensive, consistently achieves markedly superior performance on \emph{global} queries. Conversely, for \emph{local} queries, the advantages of long-context solutions diminish significantly. Notably, in the news dataset, VanillaRAG outperforms its counterpart across all four evaluation metrics. This aligns with our intuition that RAG is particularly advantageous for extracting fine-grained information needed for \emph{local} queries, whereas long-context solutions excel in addressing \emph{global} queries that demand a comprehensive understanding of memory. The above results indicate that, similar to the findings of AI-native memory ~\citep{shang2024ai}, although RAG and long-context LLMs can access the correct answer within the provided context, they do not always produce the correct response.

\section{Our \our}

Building on our findings, we propose a novel LLM-native approach named \our, which fine-tunes a relatively small (e.g., 7B) LLM using the data synthesized following RAG principles, thereby harnessing the strengths of both RAG and long-context solutions.
In this section, we first provide an overview of our approach, followed by a detailed exposition of the \emph{global} and \emph{local} data synthesis processes, as well as the fine-tuning stage of the language model.

\subsection{Overview}
As illustrated in Figure \ref{fig:pipeline}, the key idea of \our is to synthesize high-quality data following RAG principles and tuning them into the LLM parameters. The data synthesis strategy is designed to ensure the final tuned model to be versatile and context-aware.

In our implementation, we particularly choose GraphRAG~\cite{edge2024local}, as it is a recent advanced RAG method capable of constructing hierarchical memory. We focus on crafting query-answer pairs from text units, entities, and relationships. Specifically, we generate data from both \emph{local} and \emph{global} perspectives:
(1) \emph{local} data synthesis concentrates on generating content-specific query and answer pairs, and (2) \emph{global} data synthesis focuses on producing query and answer pairs that integrate insights across entities and relationships.

Table~\ref{tbl:graph_rag_stats} presents detailed statistics of the synthesized data, offering insights into the graph structure constructed by GraphRAG, including the number of entities, relations, and communities. Additionally, Table~\ref{tbl:graph_rag_stats} also summarizes the synthesized SFT Data, detailing the number of queries, average query token count, and average answer token count.
With the synthesized data, fine-tuning the LLM becomes a natural progression, where we utilize the widely adopted LoRA technique~\cite{hu2021lora}. 

Next, we will delve into the details of each component of the proposed \our method, namely the \emph{local} and \emph{global} data synthesis strategy, as well as the fine-tuning process for the LLM.

\subsection{Global Data Synthesis}
Building upon the GraphRAG constructed graph, the \emph{global} data synthesis process can be divided into two parts, based on the graph components used, namely entity-based data synthesis and relationship-based data synthesis.

\paragraph{Entity-based Data Synthesis.}
For each entity, we craft a description using meticulously designed templates tailored to the entity type, such as a person, event, or object. These templates facilitate the creation of natural and engaging questions, prompting the model to examine the role of the entity within a broader context during the subsequent query-and-answer pair generation phase.
In practice, to ensure detailed and coherent answers, we adopt the chain-of-thought (CoT) reasoning framework~\cite{wei2022chain}, resulting in more comprehensive and accurate responses. Specifically, the approach comprises the following three key steps:

\begin{enumerate}[nosep,leftmargin=*]
    \item \textbf{Restating the context}: Commence the response by concisely summarizing the situation or topic, ensuring a seamless flow and clarity, so that the answer remains coherent and contextually grounded.
    \item \textbf{Integrating entity description}: Merge essential details about the entity with pertinent information from the broader context, crafting a more nuanced and insightful answer that adds depth and relevance.
    \item \textbf{Constructing a detailed answer}: Offer a thorough and detailed explanation, typically ranging from 300 to 500 words, to comprehensively address the query, making use of all the available relevant information.
\end{enumerate}

Moreover, to enhance clarity, we employ subheadings and bullet points to organize the content. This structured approach ensures that the generated questions and answers effectively capture both specific details and the broader context. 

\begin{figure}[t]
    \centering
    \includegraphics[width=\linewidth]{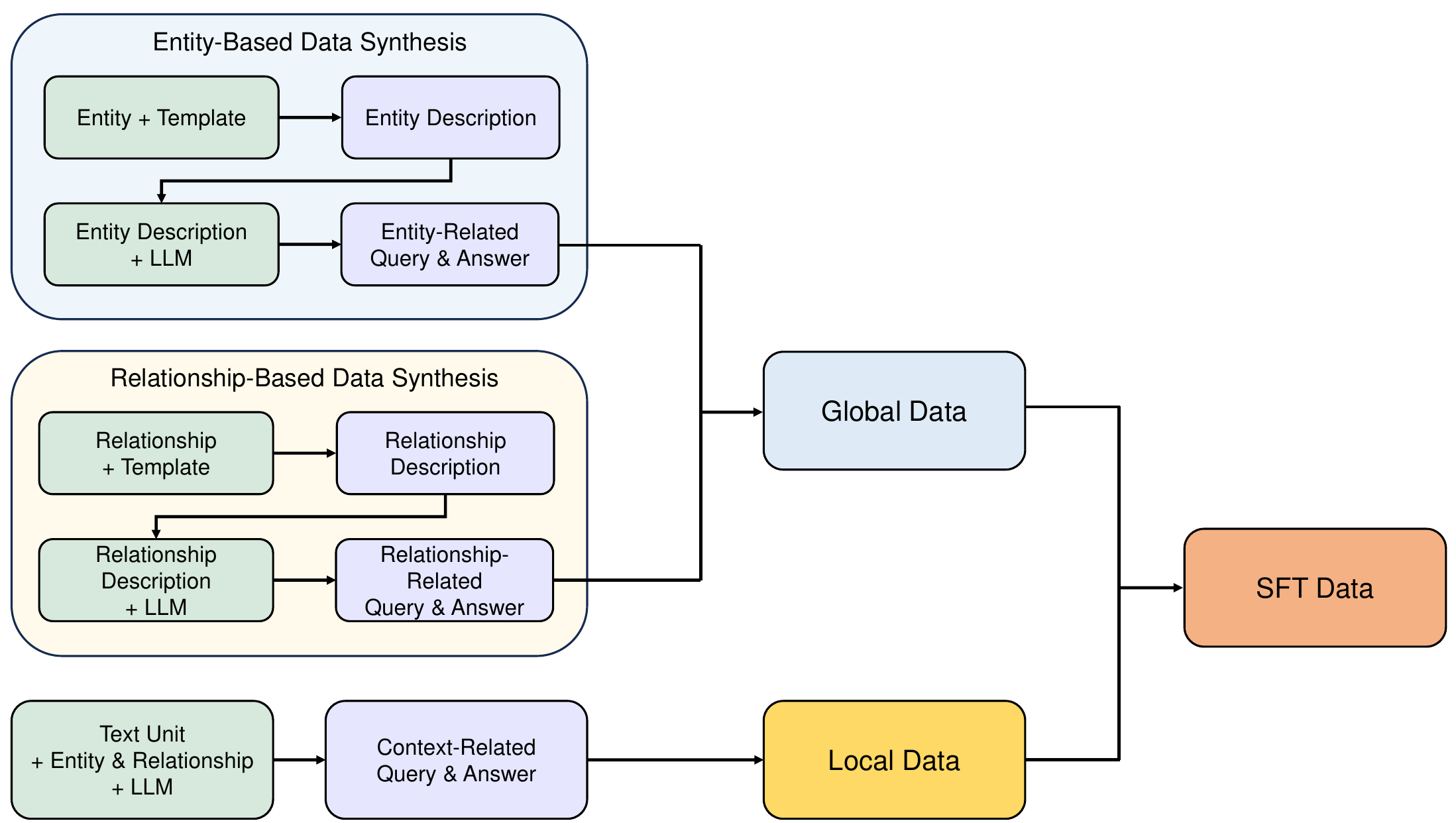}
   \caption{Overview of the data synthesis process used in \our. \emph{Global} data synthesis comprises entity-based and relationship-based data synthesis, which generates query-answer pairs through the integration of templates and LLMs. \emph{Local} data synthesis generates query-answer pairs using text units enriched by entries and relationships, along with LLMs.}
    \label{fig: data_synthisis}
\end{figure}

\paragraph{Relationship-based Data Synthesis.}
In a manner similar to entity-based data synthesis, we utilize relationship-specific templates to generate queries that delve into how entities interact. 
By merging entity and relationship-based queries with CoT reasoning-generated answers, the model can better understand both detailed insights and broader perspectives. Figure \ref{fig: data_synthisis} depicts the overall \emph{global} data synthesis process.

\subsection{Local Data Synthesis}

\emph{Local} data synthesis involves generating queries from text units that encompass multiple entities and relationships, with an emphasis on \emph{local} details. These text units offer the context needed to craft queries that investigate specific, localized aspects of the entities or relationships. The process includes:

\begin{enumerate}[nosep,leftmargin=*]
    \item \textbf{Assessing \emph{local} information}: The text units is examined to identify the pertinent entities or relationships, concentrating on the specific details within the given context.
    \item \textbf{Generating context-specific queries}: Queries are crafted based on the roles of these entities or relationships within the localized context, using the text units as the immediate reference.
\end{enumerate}

These localized queries focus on specific interactions or characteristics within the text, providing detailed insights into the smaller components of the data. As Figure \ref{fig: data_synthisis} shows, integrating \emph{local} and \emph{global} data produces the final SFT dataset, with the entire data synthesis process adhering to RAG principles.

\begin{table*}[t]
\centering
\caption{Winning rates (averaged across four evaluation metrics) of our \textbf{\our} compared to VanillaRAG, GraphRAG, Long-context LLM, and Normal SFT on the Podcast, News, and Journaling datasets. Local and Global refer to different evaluation contexts. For comparison, the check mark indicates the characteristics employed by each method. Winning rates exceeding \textbf{50\%} confirm that our \our outperforms all the compared methods.}
\resizebox{\textwidth}{!}{
\begin{tabular}{ccccccccccccc}
\toprule
\multicolumn{4}{c}{Methods}                      &  & \multicolumn{2}{c}{Podcast} &  & \multicolumn{2}{c}{News} &  & \multicolumn{2}{c}{Journaling} \\
Type & RAG Principle & LLM-Native & Parameterized Memory &  & Local        & Global       &  & Local      & Global      &  & Local         & Global         \\ 
\cmidrule(lr){1-4} \cmidrule(lr){6-7} \cmidrule(lr){9-10} \cmidrule(lr){12-13} 
VanillaRAG           & \CheckmarkBold & \XSolidBrush & \XSolidBrush &  & 94.80\%  & 96.20\%  &  & 94.60\%  & 95.80\%  &  & 81.67\%  & 95.56\%  \\
Long-context LLM     & \XSolidBrush & \CheckmarkBold & \XSolidBrush &  & 65.60\%  & 67.60\%  &  & 94.00\%  & 95.60\%  &  & 66.67\%  & 73.33\%  \\
Normal SFT           & \XSolidBrush & \CheckmarkBold & \CheckmarkBold &  & 100.00\% & 100.00\% &  & 100.00\% & 100.00\% &  & 100.00\% & 100.00\% \\
Averaged GraphRAG    & \CheckmarkBold & \XSolidBrush & \XSolidBrush &  & 57.95\%  & 57.95\%  &  & 56.35\%  & 57.41\%  &  & 51.67\%  & 59.31\%  \\
\our (Ours) & \CheckmarkBold & \CheckmarkBold & \CheckmarkBold &  & N/A    & N/A    &  & N/A    & N/A    &  & N/A    & N/A    \\ \bottomrule
\end{tabular}
}
\label{tab:rag-tuned-llm beats all}
\end{table*}

\subsection{LM tuning}
The combination of entity-based, relationship-based, and localized context-based query-answer pair generation facilitates fine-tuning an LLM to natively embody the memory extracted through GraphRAG, i.e., LLM-native memory, thereby combining the strengths of both RAG and LLM-native solutions (e.g., long-context LLMs).

While full fine-tuning \citep{lv2023full} generally achieves a higher performance ceiling, it demands significantly more computational resources and extensive training data. Furthermore, full fine-tuning may compromise the base model's generalization ability. Given the relatively small-scale fine-tuning data, we adopt LoRA, a widely used PEFT \citep{ding2023parameter} method, to parameterize a base LLM with the memory generated via RAG methods.

\section{Experiments}

\subsection{Experimental Setup}

\paragraph{Datasets and Evaluation Metrics.}
We consider the three datasets introduced in Section 2, namely News, Podcast, and Journaling. Detailed statistics and characteristics of these datasets are provided in Table\ref{tbl:dataset_stats}. Evaluation metrics are also in consistent with the four introduced in Section 2, namely helpful, rich, insightful and user-friendly.

\paragraph{Compared Methods.}
To investigate the superiority of our proposed \our, we compare it with other four methods, i.e., \textbf{VanillaRAG}, \textbf{GraphRAG}, \textbf{Long-Context LLM}, and \textbf{Normal SFT}. For VanillaRAG and the long-context LLM, we adopt the configurations detailed in Section 2, utilizing \texttt{GPT-4o-mini} with plain documents as external memory for VanillaRAG and \texttt{Gemini-1.5-pro-001} for the long-context LLM. GraphRAG is a recently advanced RAG technique, which can generate responses leveraging four hierarchical graph community information integration strategies, ranging from high-level to fine-grained, labeled \textbf{C0 to C3}:
\begin{itemize}[nosep,leftmargin=*]
    \item \textbf{C0} employs root-level community summaries.
    \item \textbf{C1} employs sub-communities of C0 but still high-level community summaries.
    \item \textbf{C2} employs intermediate-level community summaries.
    \item \textbf{C3} employs low-level community summaries.
\end{itemize}
The language model for GraphRAG is also set to \texttt{GPT-4o-mini}. For the normal SFT method, we follow ~\cite{Jiang2024InstructiontunedLM} to transform raw data into query-answer pairs for finetune an LLM, adopting the same setting as \our, i,e., selecting \texttt{Qwen-2-7B-instruct}~\cite{qwen2} as the base model, and employ a LoRA with its rank $r=64$ for parameterizing the model's memory. It is important to note that all methods are fundamentally provided with the same dataset, albeit processed in different formats.

\paragraph{Training Configurations.}
In the training process, we adopt a cosine learning rate scheduler, with a maximum learning rate of $1\times 10^{-4}$, and set the total number of fine-tuning epochs to $3$. To ensure more stable results, we set the decoding temperature to $0$ during inference.

\subsection{Superiority of \our}
Table \ref{tab:rag-tuned-llm beats all} summarize the winning rate of our proposed \our against other four compared methods. Our key point is that \our can effectively handles both \emph{local} and \emph{global} queries simultaneously, while others can not. Therefore, we report the average result across four evaluation metrics and focus on the overall result regarding different query types. Moreover, for simplity of our interpretation and comparison, we also average the results of four different GraphRAG levels, i.e., C0 to C3, and you can refer to Table 6 and 7 in Appendix for detailed results.

From the results, it is evident that our \our outperforms all competitors in addressing both \emph{local} and \emph{global} problems, with its superiority being particularly pronounced when compared to VanillaRAG, long-context LLM, and Normal SFT. We attribute the success to the fact that the RAG data enables the model to obtain fine-grained factual information for the problem, while the tuning of the memory to be LLM-native provides a deeper, more \emph{global} understanding of the issue. Furthermore, from the comparison with normal SFT, we can find that though given the same external memory, the formulation of the training data synthesis has a great influence on the model performance. GraphRAG emerges as the most competitive baseline, likely due to its incorporation of both fine-grained and high-level information in its responses. The graph it generates includes both abstract and varied levels of information, while the RAG approach retains the advantage of relevant information integration when generating responses. However, GraphRAG still inherits the conventional limitation of RAG, relying on external data sources for its responses. We argue that parameterizing the memory to be LLM-native is more effective than retrieval-based approaches. By integrating relevant information directly into the model's parameters, the LLM can generate more coherent and contextually aware responses without the need to repeatedly access external sources, ultimately improving both the efficiency and quality of the answers.

\begin{figure}
    \centering
    \includegraphics[width=\linewidth]{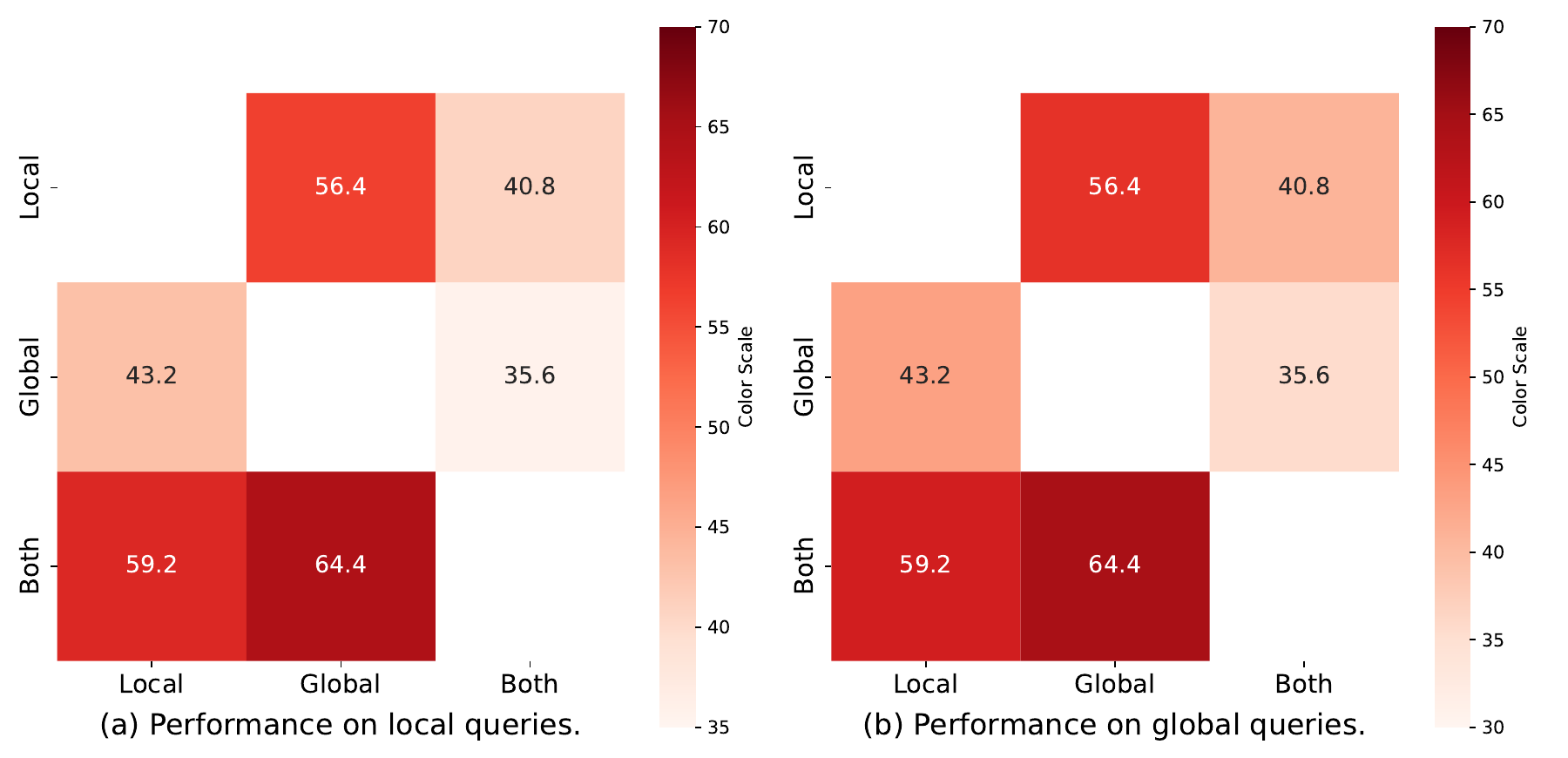}
    \caption{The comparison among \our models trained with different synthesized data types, i.e., \emph{local} split, \emph{global} split, and both. We evaluate the models on local and \emph{global} queries separately to ablate the effect of training data.}
    \label{fig:heatmap}
\end{figure}

\subsection{Ablation Studies on the Training Data}
Recall that our goal is for \our to excel at both \emph{local} and \emph{global} queries. Therefore, our data synthesis strategy also consists of two parts: \emph{local} and \emph{global} data synthesis. In this section, we will investigate how the type of training data influences the model's performance. Specifically, we consider three scenarios in the Podcast transcripts dataset: LLM tuning with \emph{local} data only, \emph{global} data only, and both \emph{local} and \emph{global} data combined. In order to better understand the effects of \emph{local} and \emph{global} data, we evaluate the tuned model separately on \emph{local} and \emph{global} queries. The winning rates of one training data type against another are illustrated in Figure \ref{fig:heatmap}.

As we can observe in the figure, models tuned with \emph{local} data perform better on \emph{local} queries than those tuned with \emph{global} data, and vice versa. When both \emph{local} and \emph{global} data are combined, the model achieves the best results on both \emph{local} and \emph{global} queries. This highlights the benefit of using diverse training data types, enhancing the model's robustness and generalization. These ablation studies also demonstrate the profound impact that training data has on the performance of a deep learning model.

\begin{table}[t]
\centering
\caption{Zero-shot performance comparison between the original base model and our \our across three distinct capabilities. }
\label{tbl:experiment_results}
\resizebox{\linewidth}{!}{
\begin{tabular}{c c c c}
\toprule
\textbf{Dataset}  & \textbf{Capability} & \textbf{Original Model} & \textbf{\our} \\ \midrule
MMLU              & English              & 80.80\%                         & 73.50\%                          \\
GSM8K             & Mathematics          & 63.66\%                        & 61.72\%                          \\
HumanEval         & Coding               & 57.90\%                         & 56.70\%                           \\ \bottomrule
\end{tabular}
}
\end{table}

\subsection{Evaluation of Generalization Capability}

Since \our can be understood as training and testing within a fixed knowledge domain, it is natural for us to evaluate the model’s generalization ability. We divide generalization into two aspects: (1) the ability to answer unseen queries within the same knowledge domain and (2) the model’s general capability beyond the given domain. For the first aspect, our test queries are generated using methods significantly different from those used for the training data, meaning that the test results inherently reflect the model’s ability to answer out-of-training-distribution queries within the domain. Therefore, the following evaluation will primarily focus on the second aspect.

To illustrate the generalization capability beyond the given domain, we compare its zero-shot performance with that of the original base model across three widely recognized large-scale benchmarks: MMLU~\citep{hendrycks2020measuring}, GSM8K~\citep{cobbe2021training}, and HumanEval~\citep{chen2021evaluating}. Specifically, we utilize the model fine-tuned on \textbf{News} articles, as it encompasses the largest volume of training tokens. The experimental results summarized in Table~\ref{tbl:experiment_results} reveal that \our incurs only a slight degradation in performance compared to the original base model, thereby underscoring its robust generalization capability.

\section{Related Works}
\subsection{Retrieval Augment Generation}
Pre-trained language models, such as Qwen \cite{bai2023qwen} and Llama \cite{touvron2023llama}, have shown impressive query-answering capabilities. However, they face limitations when tasked with problems requiring knowledge beyond their training data. Retrieval-augmented generation (RAG) \cite{lewis2020retrieval} provides a solution by retrieving relevant information from an external knowledge base. While RAG has proven to be practical and effective, traditional RAG systems can only retrieve raw corpus related to the query, without broader comprehension. As a result, abstract queries such as those asking for high-level insights or overarching understandings often lead to suboptimal answers. To overcome these limitations, GraphRAG \cite{edge2024local} has been introduced. Specifically, GraphRAG constructs a knowledge graph using an LLM, enabling it to provide hierarchical information that range from specific, detailed facts to more global, abstract insights, leveraging the knowledge graph for a more comprehensive understanding

\subsection{Long-context LLM}
Long-context LLMs are designed to handle tasks that involve processing extended sequences of text, addressing a significant limitation of traditional LLMs, which typically operate with fixed, limited context windows. For example, GPT-4o~\citep{achiam2023gpt} offers a context window of up to 128K tokens, while Gemini 1.5~\citep{reid2024gemini} can manage up to 1M or 10M tokens. Furthermore, various studies have sought to push the boundaries of these context windows, suggesting models capable of "unlimited" context lengths through innovations such as length extrapolation~\citep{peng2023yarn, xiao2023efficient, han2023lm, zhang2024soaring} and position bias adjustments~\citep{liu2024lost, peysakhovich2023attention, an2024make}. Long-context LLMs, in principle, possess the potential to offer more refined abstraction abilities and a deeper, more nuanced understanding of global context compared to RAG methods. Yet, as highlighted by \citet{hsieh2024ruler, shang2024ai}, the context may surpass the constraints of the LLM’s context window, which is typically much narrower than reported, leading to the inadvertent loss of crucial information amid an expansive sea of text.

\subsection{Fine-Tuning LLMs}
To incrementally expand the knowledge of a pre-trained LLM or to align it with human preferences, fine-tuning stands as one of the most prevalent approaches, encompassing methods such as supervised fine-tuning (SFT), reinforcement learning from human feedback (RLHF) \cite{ouyang2022training}, and direct preference optimization (DPO) \cite{rafailov2024direct}. Despite their effectiveness, these techniques are notably annotation-heavy and computationally intensive, rendering the fine-tuning of an LLM using these methods prohibitively costly. To circumvent the extensive computational demands of full fine-tuning, which can reach into tens of billions, numerous parameter-efficient fine-tuning (PEFT) methods have been explored, including BitFit \cite{zaken2021bitfit}, adapter \cite{houlsby2019parameter}, and Lora \cite{hu2021lora}. In this paper, we primarily employ a LoRA to fine-tune a \our. Methodologically, RAFT \citep{zhang2024raftadaptinglanguagemodel} is the closest to our approach, as it explores the potential integration of RAG and fine-tuning. However, there are two fundamental differences between our work and RAFT: First, the model we train is not intended for use in the generation stage of RAG, making our objectives fundamentally different; Second, our training data does not include deliberately introduced noise, which distinguishes our approach significantly in terms of methodology.

\section{Conclusion and Future Work}
In this paper, we validate RAG's fine-grained retrieval abilities and the global abstraction strengths of LLM-native solutions. However, RAG lacks holistic understanding, and long-context models tend to lose key information over extended contexts. We integrate these strengths of both RAG and LLM-native solutions by fine-tuning an LLM within an RAG framework for data generation. This work is the first to explore LLM and RAG integration within a unified framework, bridging open-domain and domain-specific query-answering tasks. Our RAG-Tuned LLM, equipped with LLM-native memory, outperforms both standard RAG methods and long-context LLMs across diverse datasets, demonstrating superior performance in handling hierarchical queries.

\paragraph{Future Work.} 
Building on this study, several future directions are worth exploring to further validate and enhance our proposed method. First, we plan to extend \our to more diverse datasets and domains, enabling us to evaluate its generalizability across different tasks, including complex challenges like multi-hop reasoning and multi-modal query-answering. This will provide a clearer understanding of \our's effectiveness in both open-domain and domain-specific contexts. Additionally, we will experiment with various foundational models (e.g., the Llama series \cite{touvron2023llama, touvron2023llama2}), evaluating \our's adaptability to different architectures and model scales. This will highlight the trade-offs between model size, computational efficiency, and performance when combining RAG and LLM-native methods.

\newpage
\section*{Limitations}

While our proposed method, \our, demonstrates substantial advantages over long-context LLMs and RAG in handling both \emph{global} and \emph{local} queries, we recognize two key limitations that warrant further investigation. First, although LLM-as-a-judge is a widely adopted evaluation approach \cite{li2024generation}, the metrics we utilized remain relatively domain-specific—suitable for applications like personal assistants but less adaptable to general-purpose language models. Enhancing the robustness and generalizability of our evaluation framework is imperative. Second, although we have validated our method’s robustness and generalization to some extent (e.g., in English, mathematics, and coding capabilities), broader exploration such as in the realms of multi-modal and multi-hop reasoning tasks remains insufficient.

\bibliography{custom}

\clearpage

\appendix

\begin{figure*}[t]
\caption{A concrete example (Case 1) from the News dataset illustrating the superiority of \our compared to GraphRAG.}
\label{fig: case1}
\centering
\begin{minipage}{\textwidth}  
\begin{tcolorbox}[width = \textwidth, title =Case 1, colback=blue!5, colframe= gray]
\textbf{Query:} What is the expected release date of Britney Spears' memoir?\\

\textbf{GraphRAG C3 Answer:} Britney Spears' highly anticipated memoir, titled \textit{The Woman in Me}, is set to be released on \textbf{October 24, 2023}. This memoir is expected to delve into significant moments in her life, including her experiences with fame, motherhood, and the impact of her conservatorship, which has been a focal point of public interest and discussion.\\

The memoir is described as a candid account of her life, providing insights into her personal struggles and triumphs, and is anticipated to resonate deeply with her fans and the general public alike.\\

\textbf{\our Answer:} Britney Spears' memoir, titled \textit{The Woman in Me}, is anticipated to be released on \textbf{October 24, 2023}. This highly anticipated book is set to delve into her life experiences and challenges, particularly focusing on her conservatorship, a defining aspect of her public persona for many years. The memoir promises to provide readers with an intimate look at her journey through fame, personal struggles, and the complexities of her relationships, including her marriage to Sam Asghari.\\

Following a tumultuous period marked by the conservatorship, which was finally terminated in November 2021, Britney's memoir comes at a pivotal moment in her life. The end of the conservatorship allowed her to regain control over her life and career, offering her a sense of newfound freedom. In interviews, she has expressed sadness about the loss of her childhood home, describing it as a place where she felt most like herself. This emotional context adds depth to her narrative, as she explores themes of identity, resilience, and the impact of fame on her personal relationships. \\

The memoir is not merely a recounting of events; it is also a reflection on her personal growth and the lessons she has learned along the way. Britney has been open about her struggles with mental health and the pressures of public life, making her story relatable to many who have faced similar challenges. The memoir is also expected to offer insights into her relationship with Sam Asghari, whom she married in June 2022. Their marriage ended in July 2023, adding another layer of complexity to her story. \\

Overall, \textit{The Woman in Me} is poised to be a significant literary work, capturing the essence of Britney Spears' life and offering readers a chance to understand her perspective on fame, motherhood, and her quest for autonomy. As the release date approaches, there is growing anticipation about how Britney will share her story and what new revelations may come to light.\\

\end{tcolorbox}
\end{minipage}
\end{figure*}

\begin{figure*}[t]
\caption{A concrete example (Case 2) from the News dataset illustrating the superiority of \our compared to GraphRAG.}
\label{fig: case2}
\centering
\begin{minipage}{\textwidth}  
\begin{tcolorbox}[width = \textwidth, title =Case2, colback=blue!5, colframe= gray]

\textbf{Query:} How many counts of fraud and money laundering is Sam Bankman-Fried (SBF) facing in his trial?\\

\textbf{GraphRAG C3 Answer:}
Sam Bankman-Fried (SBF) is facing a total of seven counts related to his trial, which include charges of fraud and conspiracy. Specifically, these charges encompass wire fraud and money laundering, among other allegations tied to the operations of FTX and the misappropriation of customer funds.\\ 

The trial has garnered significant attention due to the serious nature of these allegations, which are considered some of the largest financial frauds in recent history.\\

\textbf{\our Answer:}
In the ongoing trial of Sam Bankman-Fried, commonly referred to as SBF, he is facing serious legal charges that include multiple counts of fraud and money laundering. The context surrounding this trial is quite significant, as it has garnered considerable attention due to the collapse of FTX, a cryptocurrency exchange that was once valued at an impressive \$32 billion before its downfall. This dramatic shift in fortune has led to a high-profile legal battle that has captured the interest of both the public and legal experts alike.\\

The trial itself is set against a backdrop of allegations that SBF engaged in fraudulent activities while managing FTX and its affiliated trading firm, Alameda Research. Specifically, the prosecution claims that SBF orchestrated a scheme that misappropriated customer funds, leading to substantial financial losses for investors and customers. The prosecution's case is expected to unfold over several weeks, with the government presenting evidence that will detail how SBF allegedly misled investors and customers about the safety and management of their assets.\\

As part of the trial proceedings, SBF has entered a plea of not guilty, which indicates his intent to contest the charges against him. The defense's strategy will likely revolve around challenging the prosecution's narrative and presenting arguments that could potentially exonerate him. However, the prosecution's case is bolstered by the testimony of key witnesses, including former associates and industry experts who can provide insight into the operations of FTX and the broader implications of the alleged fraud.\\

In summary, Sam Bankman-Fried is facing multiple counts of fraud and money laundering in his trial, which is a critical juncture for both him and the cryptocurrency industry as a whole. The outcome of this trial could have far-reaching consequences, not only for SBF but also for regulatory frameworks governing cryptocurrency exchanges moving forward. As the trial progresses, it will be essential to monitor how the evidence presented impacts the jury's perception and ultimately influences the verdict.

\end{tcolorbox}
\end{minipage}
\end{figure*}

\section{Definition of Global and Local Queries}
\label{appendix: A}
A notable innovation in our query generation method lies in the differentiation between \emph{global} and \emph{local} queries, akin to the approach used in GraphRAG, but with a more pronounced emphasis on user-driven tasks. Particularly, we define \emph{local} and \emph{global} queries as follows:
\begin{itemize}[nosep,leftmargin=*]
    \item \textbf{Global Queries}: Global queries are crafted to elicit high-level, interpretive responses that require the user to consider the dataset in its entirety. They address overarching trends, themes, and insights that emerge from the data, steering the user toward macro-level analysis. Therefore, global query synthesis demands multiple dataset chunks, ensuring that the user engages with the dataset holistically, rather than fixating on specific details.
    
    \item \textbf{Local Queries}: Local queries are retrieval-oriented, aiming to direct the user toward specific pieces of information within the dataset. Each query is designed to be answerable by referencing a particular section or chunk of the data, promoting a detailed and focused analysis. Local queries necessitate precision in information retrieval and cater to users seeking clear, concrete answers to more narrowly defined questions.
\end{itemize}
By categorizing the queries into these two types, we ensure that the evaluation of RAG systems encompasses both granular detail retrieval and broader sensemaking tasks, thereby offering a more comprehensive assessment of the system's capability to engage with the dataset at multiple levels.

\section{Explaination of Evaluation Metrics} \label{appendix: metric}
\label{appendix: B}

\begin{itemize}[nosep, leftmargin=*]
    \item \textbf{Helpful}: This metric evaluates the accuracy and reliability of the answer in relation to the posed query. It examines whether the answer directly addresses the query and delivers useful, relevant information. Answers that exhibit clear correctness and offer valuable content receive higher scores on this metric.
    
    \item \textbf{Rich}: This metric evaluates the variety and depth of the content provided in the answer. An answer that explores multiple perspectives or offers detailed explanations from different angles is deemed more diverse and rich. It emphasizes comprehensiveness and the ability to present a nuanced understanding of the dataset or topic.
    
    \item \textbf{Insightful}: This metric measures the depth of understanding demonstrated in the answer. Insightful responses reflect a profound comprehension of the subject matter and may offer thoughtful or original insights that transcend surface-level retrieval. Answers that meaningfully synthesize data to provide novel or perceptive interpretations receive higher ratings.
    
    \item \textbf{User-Friendly}: This metric assesses the clarity, readability, and organization of the response. An answer that is well-structured, concise, and easily comprehensible will score higher. This metric ensures that even complex responses remain accessible and understandable to the target audience, striking a balance between depth and usability.
\end{itemize}

\section{Results of Local and Global Subsets}
\label{appendix: C}

Table \ref{tab:rag-tuned-llm beats all} in the main body of the paper only summarizes the averaged results across four evaluation metrics and four distinct levels of GraphRAG responses. In this section, we provide more detailed results for each metric and each level of GraphRAG responses. Table \ref{tab:local_results} and \ref{tab:global_results} shows the winning rates of our \our over GraphRAG (C0 to C3), Long-context LLM, VanillaRAG, and normal SFT on \emph{local} and \emph{global} queries, respectively. The results demonstrate that our \our generally outperforms all the compared methods across all metrics.

\section{Exampls of \our vs. GraphRAG}
\label{appendix: D}

As shown in Table \ref{tab:local_results} and \ref{tab:global_results}, GraphRAG is the strongest competitor among the four methods compared. Therefore, we present two concrete examples to qualitatively demonstrate the superiority of \our over GraphRAG, beyond numerical performance, as shown in Figure \ref{fig: case1} and \ref{fig: case2}.



\begin{table*}[t]
\centering
\caption{Winning rates (\%) of our \our over GraphRAG (C0 to C3), Long-context LLM, VanillaRAG, and Normal SFT across four evaluation metrics on \emph{local} queries.}
\resizebox{\textwidth}{!}{\begin{tabular}{@{}ccccccccc@{}}
\toprule
Dataset                  & Metric            & GraphRAG C0 & GraphRAG C1 & GraphRAG C2 & GraphRAG C3 & Long-Context LLM & VanillaRAG & Normal SFT \\ \midrule[0.3mm]
\multirow{4}{*}{Podcast} & Helpful        & 56.80        & 53.60        & 52.00        & 52.80     & 65.60   & 95.20 & 100.00     \\
                         & Rich          & 52.80        & 49.60        & 47.20        & 48.00   & 59.20  & 96.00  & 100.00    \\
                         & Insightful    & 59.20        & 54.40        & 50.40        & 51.20        & 60.00  & 99.20  & 100.00    \\
                         & User-Friendly & 80.00        & 76.00        & 72.00        & 71.20        & 77.60    & 88.80  & 100.00    \\ \midrule
\multirow{4}{*}{News}    & Helpful        & 52.00        & 52.80        & 49.60        & 50.40    & 95.20     & 95.20   & 100.00  \\
                         & Rich          & 50.40        & 49.60        & 45.60        & 46.40   & 94.40  & 99.20  & 100.00    \\
                         & Insightful    & 56.00        & 55.20        & 51.20        & 51.20   & 96.00    & 99.20   & 100.00   \\
                         & User-Friendly & 78.40        & 73.60        & 70.40        & 68.80     & 90.40  & 84.80   & 100.00   \\ \midrule
\multirow{4}{*}{LPM}     & Helpful        & 53.33        & 46.67        & 46.67        & 46.67   & 60.00   & 73.33 & 100.00     \\
                         & Rich          & 46.67        & 53.33        & 46.67        & 46.67     & 66.67  & 86.67 & 100.00      \\
                         & Insightful    & 66.67        & 60.00        & 53.33        & 60.00        & 73.33  & 86.67   & 100.00   \\
                         & User-Friendly & 53.33        & 46.67        & 53.33        & 46.67   & 66.67  & 80.00 & 100.00      \\ \bottomrule
\end{tabular}}
\label{tab:local_results}
\end{table*}

\begin{table*}[t]
\centering
\caption{Winning rates (\%) of our \our over GraphRAG (C0 to C3), Long-context LLM, VanillaRAG, and Normal SFT across four evaluation metrics on \emph{global} queries.}
\resizebox{\textwidth}{!}{\begin{tabular}{@{}ccccccccc@{}}
\toprule
Dataset                  & Metric            & GraphRAG C0 & GraphRAG C1 & GraphRAG C2 & GraphRAG C3 & Long-Context LLM & VanillaRAG & Normal SFT\\ \midrule[0.3mm]
\multirow{4}{*}{Podcast} & Helpful        & 54.40        & 55.20        & 52.80        & 52.00    & 68.00  & 97.60  & 100.00    \\
                         & Rich          & 52.00        & 49.60        & 47.20        & 45.60        & 61.60  & 97.60 & 100.00     \\
                         & Insightful    & 59.20        & 52.80        & 53.60        & 50.40    & 62.40   & 99.20 & 100.00      \\
                         & User-Friendly & 82.40        & 76.80        & 73.60        & 69.60        & 78.40   & 90.40 & 100.00     \\ \midrule
\multirow{4}{*}{News}    & Helpful        & 52.80        & 53.60        & 51.20        & 52.00    & 96.80    & 98.40  & 100.00    \\
                         & Rich          & 51.20        & 48.80        & 46.40        & 46.60    & 96.00      & 99.20  & 100.00    \\
                         & Insightful    & 56.80        & 56.00        & 52.00        & 52.00        & 97.60     & 100.00 & 100.00    \\
                         & User-Friendly & 79.20        & 73.60        & 71.20        & 68.80        & 92.00    & 85.60  & 100.00    \\ \midrule
\multirow{4}{*}{LPM}     & Helpful        & 57.78        & 55.56        & 55.56        & 53.33        & 64.44   & 93.33  & 100.00    \\
                         & Rich          & 55.56        & 57.78        & 57.78        & 55.56        & 73.33   & 100.00  & 100.00   \\
                         & Insightful   & 68.89        & 68.89        & 64.44        & 68.89        & 82.22   & 100.00 & 100.00    \\
                         & User-Friendly & 57.78        & 57.78        & 57.78        & 55.56        & 73.33    & 88.89 & 100.00     \\ \bottomrule[0.5mm]
\end{tabular}}
\label{tab:global_results}
\end{table*}

\end{document}